\documentclass[10pt,twocolumn,letterpaper]{article}
\usepackage{makecell}

\usepackage[pagenumbers]{iccv} 
\usepackage{xcolor}
\usepackage{colortbl}
\usepackage{amsmath, amssymb, amsthm}  
\usepackage{algorithm}
\usepackage{algpseudocode}

\newtheorem{theorem}{Theorem}  
\definecolor{iccvblue}{rgb}{0.21,0.49,0.74}
\usepackage[pagebackref,breaklinks,colorlinks,allcolors=iccvblue]{hyperref}
\usepackage[utf8]{inputenc}
\usepackage{multirow} 
\usepackage{algorithm}
\def\eg{\emph{e.g.}}

\title{Lie Detector: Unified Backdoor Detection via Cross-Examination Framework}

\author{Xuan Wang\\
{\tt\small wangxuan21d@nudt.edu.cn}
\and
Siyuan Liang\\
{\tt\small siyuan96@nus.edu.sg}
\and
Dongping Liao\\
{\tt\small	yb97428@um.edu.mo}
\and
Han Fang\\
{\tt\small 	fanghan@nus.edu.sg}
\and
Aishan Liu\\
{\tt\small liuaishan@buaa.edu.cn}
\and
Xiaochun Cao\\
{\tt\small caoxiaochun@mail.sysu.edu.cn}
\and
Yuliang Lu\\
{\tt\small publicluyl@126.com}
\and
Chang Ee-Chien\\
{\tt\small changec@comp.nus.edu.sg}
\and
Xitong Gao\\
{\tt\small xt.gao@siat.ac.cn}
}

\begin{document}
\maketitle
\begin{abstract}
Institutions with limited data and computing resources often outsource model training to third-party providers in a semi-honest setting, assuming adherence to prescribed training protocols with pre-defined learning paradigm (e.g., supervised or semi-supervised learning). However, this practice can introduce severe security risks, as adversaries may poison the training data to embed backdoors into the resulting model. Existing detection approaches predominantly rely on statistical analyses, which often fail to maintain universally accurate detection accuracy across different learning paradigms. To address this challenge, we propose a unified backdoor detection framework in the semi-honest setting that exploits cross-examination of model inconsistencies between two independent service providers. Specifically, we integrate central kernel alignment to enable robust feature similarity measurements across different model architectures and learning paradigms, thereby facilitating precise recovery and identification of backdoor triggers. We further introduce backdoor fine-tuning sensitivity analysis to distinguish backdoor triggers from adversarial perturbations, substantially reducing false positives. Extensive experiments demonstrate that our method achieves superior detection performance, improving accuracy by 5.4\%, 1.6\%, and 11.9\% over SoTA baselines across supervised, semi-supervised, and autoregressive learning tasks, respectively. Notably, it is the first to effectively detect backdoors in multimodal large language models, further highlighting its broad applicability and advancing secure deep learning. 
\end{abstract}    
\section{Introduction}

Deep learning models have grown exponentially in size in recent years, outstripping the computational resources available to many small and medium-sized institutions. Consequently, these institutions often rely on third-party cloud providers for model training. Although these providers are considered ``semi-honest'' in that they ostensibly adhere to prescribed protocols, they may still covertly manipulate data or models. This scenario can give rise to a significant \emph{backdoor threat}, where hidden triggers are embedded during training, enabling the model to function normally under most conditions but exhibit malicious behavior when specific triggers are activated \cite{badnets_2017,liu2025elba,blended_2017,ISSBA_2021,liang2023badclip, liang2024poisoned, liu2023pre, liang2024revisiting, zhu2024breaking}.

Current backdoor detection methods frequently rely on model behavior and statistical analyses (\eg, gradient-based detection, posterior analysis) \cite{NC_2019,ABS_2019,NAD_2021,DECREE_2023,MM-BD_2024,TED_2024,seer_2024, wang2022universal}. However, such approaches tend to be highly sensitive to variations in optimization objectives, loss functions, and feature representations across different learning paradigms~\cite{dargan2020survey}. This limitation constrains their ability to generalize across diverse architectures and attack strategies~\cite{liu2024compromising, zhang2024towards,xiao2024bdefects4nn,liang2024red}, posing serious challenges for maintaining user model security in a semi-honest setting.

\begin{figure}[htbp]
    \centering
    \includegraphics[width=0.5
    \textwidth]{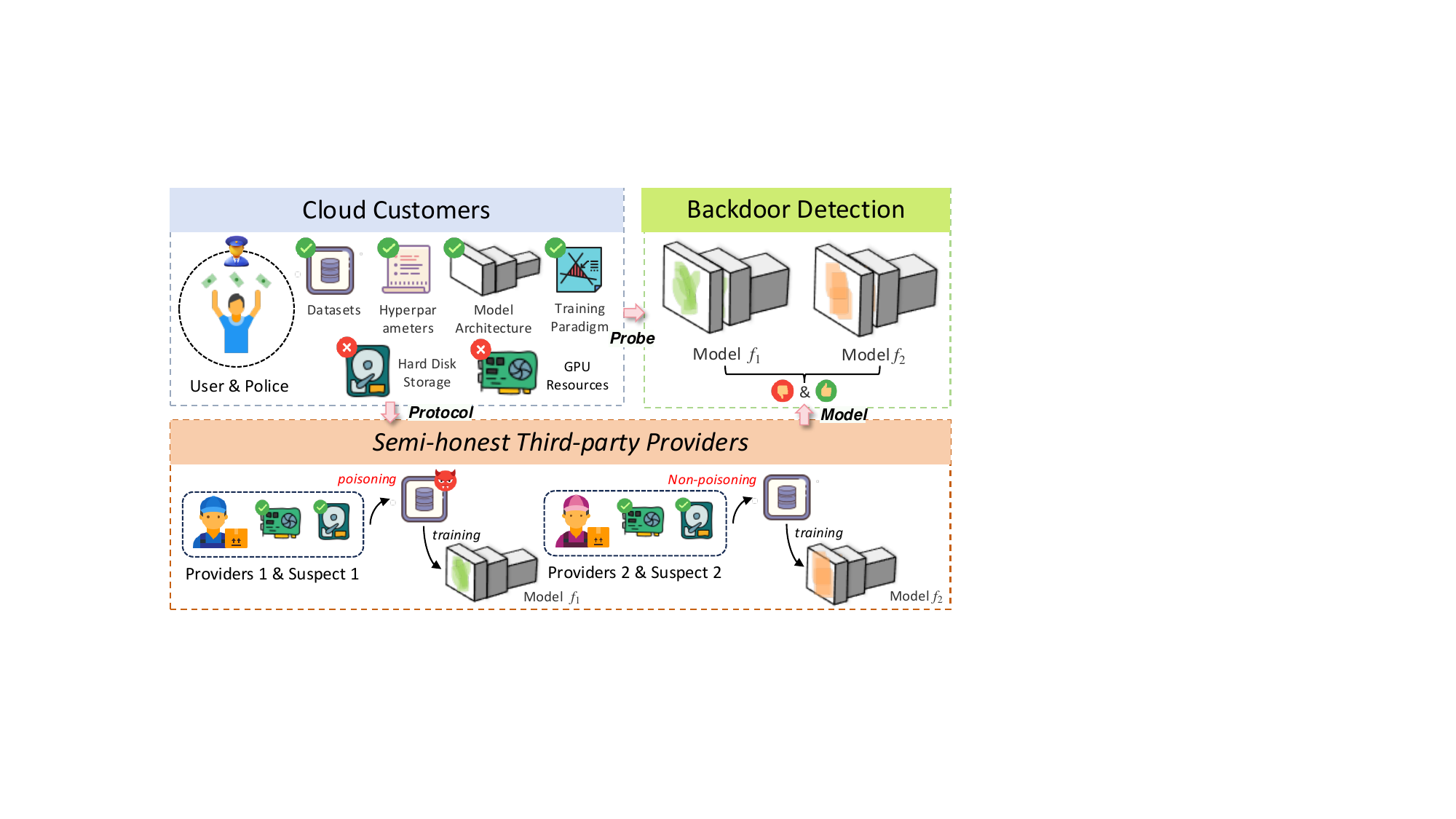} 
    \vspace{-0.3cm}
    \caption{In the absence of training resources, the user delegates model training to a third-party vendor in a semi-honest environment and generates two independent models. At the same time, the user doubles as a police to identify potential backdoor models through comparative analysis.} 
    
    \label{fig:homepage} 
    \vspace{-0.6cm}
\end{figure}

To address these shortcomings, we propose \emph{Lie Detector}, a cross-examination backdoor detection framework designed for third-party verification. As illustrated in Fig.~\ref{fig:homepage}, the user (acting as \texttt{police}) outsources the same task to two independent providers (the \texttt{suspects}) and uncovers backdoors by identifying inconsistencies in their model outputs (the \texttt{lies}). Specifically, we employ Central Kernel Alignment (CKA) \cite{cka_ori,cka_2024} for task sensitivity analysis, enabling the reverse-engineering of triggers (the \texttt{evidence}) by maximizing representational differences between clean and backdoored models. In contrast to conventional methods that depend on decision boundaries, our approach optimizes triggers based on output distributions, allowing it to generalize across supervised, semi-supervised, and autoregressive learning tasks. Additionally, we introduce a fine-tuning sensitivity analysis to distinguish truly backdoored models from benign ones, thereby reducing false positives and enhancing detection robustness. This unified framework consistently achieves high detection accuracy across multiple learning paradigms, offering a practical and versatile solution for secure backdoored model verification.

We extensively evaluate the effectiveness of \emph{Lie Detector} across supervised, semi-supervised, and autoregressive learning paradigms. The results show that our method significantly outperforms state-of-the-art backdoor detection approaches, with relative improvements of +5.4\%, +1.6\%, and +11.9\%, respectively. In addition, Lie Detector demonstrates high stability under varying random seeds, underscoring its robustness. We anticipate that this research will encourage broader adoption of secure training practices in third-party services, thereby strengthening security guarantees for deep learning models. Our \textbf{contributions} are
\begin{itemize}
    \item We design a unified cross-examination framework for backdoor detection by analyzing inconsistencies in models provided by multiple third-party service providers, enhancing the security of outsourced training in semi-honest environments.

    \item Our method combines CKA task sensitivity analysis and output distribution optimization, breaking the reliance on decision boundaries and enabling backdoor detection to generalize beyond supervised learning to semi-supervised learning and autoregressive learning.
    
    \item We achieve superior generalization, improving detection by 5.4\%, 1.6\%, and 11.9\% across three learning paradigms and seven attack methods. Notably, it is the first to enable backdoor detection in multi-modal large language models, further broadening its applicability.
\end{itemize}

\section{Related Work}
\label{sec:relatedwork}
\subsection{Development of Learning Paradigms}

Deep learning has evolved through various training paradigms to address different challenges and data types. This article focuses on supervised learning, self-supervised learning, and autoregressive learning, highlighting their motivations, advancements, and limitations.

Supervised Learning (SL) trains models on labeled data, with early breakthroughs like CNNs \cite{CNN_1998} for image classification and DNNs for speech recognition. Large-scale datasets (e.g., ImageNet \cite{ImageNet_2009}) and architectures (e.g., ResNet \cite{Resnet_2016}, VGG \cite{VGG_2015}) further advanced the field. However, its reliance on labeled data, which is costly and time-consuming to obtain, motivated the development of alternative paradigms.

Self-Supervised Learning (SSL) emerged to address the data labeling bottleneck by generating labels automatically from unlabeled data. Transformers like BERT \cite{bert_2019} revolutionized NLP, while contrastive learning frameworks like SimCLR \cite{sim_2020} excelled in vision tasks. SSL bridges the gap between supervised and unsupervised learning by leveraging inherent data structures. Contrastive Learning (CL), a subset of SSL, explicitly contrasts positive and negative samples to learn meaningful representations. Recent advancements like CLIP \cite{clip_2021} and CoCoOp \cite{coop_2022} highlight CL's versatility in unimodal and multimodal settings.

Autoregressive Learning (AL) extends SSL and CL by modeling data distributions and generating new samples across modalities. Transformer-based models like MiniGPT-4 \cite{gpt4_2023} and LLaVA \cite{llava_2023} enable joint text-image representations, advancing cross-modal understanding and generation.

The evolution from SL to SSL and AL addresses challenges in data, annotation, and generalization, enhancing model adaptability. This shift, driven by large-scale pre-trained models, has fueled deep learning advancements. However, their high computational demands limit accessibility for many users.

\subsection{Backdoor Attack}
Backdoor attacks have emerged as a critical security concern in deep learning, with their methods evolving alongside advancements in learning paradigms. These attacks aim to embed malicious behaviors into models during training, which can be triggered during inference by specific inputs. 

Early backdoor attacks primarily focused on models trained with SL, leveraging labeled datasets to embed triggers. Notable examples include BadNets \cite{badnets_2017}, which introduces poisoned data with predefined triggers to manipulate model predictions; Blended \cite{blended_2017}, which uses blended patterns as triggers, making them less detectable; ISSBA \cite{ISSBA_2021}, which embeds invisible, sample-specific triggers to enhance stealth; WaNet \cite{wanet_2021}, which utilizes warping-based triggers to achieve high attack success rates; and Low-Frequency \cite{low_2021}, which exploits low-frequency components in images to embed triggers. 

As SSL gained traction, attackers adapted existing methods and developed new techniques to target these models, which often rely on unlabeled data. Examples include BadCLIP \cite{badclip_2024}, which extends backdoor attacks to contrastive language-image pretraining models, compromising multimodal representations, and BadEncoder \cite{badencoder_2022}, which poisons the encoder in SSL frameworks, affecting downstream tasks. With the rise of generative and multimodal models, backdoor attacks have expanded to exploit the AL paradigms, such as TrojanVLM \cite{trojanvlm_2024}, which targets vision-language models by embedding triggers in multimodal data, and Shadowcast \cite{sw_2024}, which focuses on stealthy backdoor attacks in generative models, particularly in text-to-image synthesis.

The landscape of backdoor attacks is extensive and continues to grow, spanning SL, SSL, and AL paradigms. While many attacks initially targeted supervised learning, they have been adaptively transferred or redesigned for self-supervised and multimodal settings. This proliferation of attacks highlights the urgent need for robust and generic defense mechanisms to safeguard deep learning systems across all learning paradigms.

\subsection{Backdoor Detection}

\textbf{Existing Backdoor Detection Methods.} Current backdoor detection methods frequently rely on model behavior and statistical analyses, such as gradient-based detection and posterior analysis \cite{NC_2019,ABS_2019,NAD_2021,DECREE_2023,MM-BD_2024,TED_2024,seer_2024}. These approaches often analyze the internal dynamics of models, such as gradients, activations, or output distributions, to identify anomalies indicative of backdoor behavior. For instance, Neural Cleanse (NC) \cite{NC_2019} proposes an anomaly detection framework to identify and mitigate backdoors by analyzing the reversibility of triggers. Similarly, ABS \cite{ABS_2019} leverages activation clustering to detect poisoned neurons, while NAD \cite{NAD_2021} employs knowledge distillation to suppress backdoor effects during model fine-tuning. More recent works, such as MM-BD \cite{MM-BD_2024},
MM-BD \cite{MM-BD_2024} designed a universal post-training backdoor detection method that identifies arbitrary backdoor patterns by analyzing the classifier’s output landscape and applying unsupervised anomaly detection. In contrast, TED \cite{TED_2024} introduced a topological evolution dynamics framework to detect backdoors by modeling deep learning systems as dynamical systems, where malicious samples exhibit distinct evolution trajectories compared to benign ones. 
Some researches have proposed backdoor detection methods for SSL and AL paradigms, such as DECREE \cite{DECREE_2023} which achieves backdoor detection by optimizing triggers, and SEER \cite{seer_2024} which introduces another information modality for backdoor detection.

Existing backdoor detection methods have made some progress within individual learning paradigms, but their scalability is limited, making it difficult to directly apply them to other learning paradigms. In the future, there is a need to develop an unified detection methods to address backdoor threats across multiple learning paradigms.
\section{Preliminary}
This section introduces the fundamental concepts and theoretical foundations required for our method, primarily including the threat model and the definition of CKA.
\subsection{Threat Model}
In our proposed \emph{cross-examination-based backdoor detection framework}, we operate under a \emph{semi-honest adversary model} tailored for third-party model verification. 

This threat model assumes that the service providers supplying the models are semi-honest, meaning they may attempt to embed backdoors into the models but will not actively interfere with the detection process itself. The adversary's goal is to introduce hidden malicious behaviors into the model, which can be triggered by specific inputs during inference, while maintaining the model's normal functionality on clean data.

\textbf{Adversarial capabilities}. The adversary, \emph{e.g.}, a malicious service provider (the \texttt{suspects}), has the capability to inject backdoors into the model during training or fine-tuning. This could involve poisoning the training data with trigger patterns or directly manipulating the model's parameters to embed malicious behavior.

\textbf{Adversarial knowledge}. The adversary may have full knowledge of the model architecture and training process but is unaware of the specific detection mechanisms employed by the verifier. This ensures that the backdoor detection framework remains robust against adaptive attacks.

\textbf{Detection constraints}. The verifier user (the \texttt{police}) has no access to the training data or process and cannot assume the availability of a clean reference model. This aligns with real-world scenarios where third parties have no visibility into the training process, treating it as a black box.

\subsection{Centered Kernel Alignment}
CKA~\cite{cka_1,cka_2024,cka_ori} can be used to measure the similarity between activations or feature representations. To compute CKA, we first input data $\mathbf{X}$ into two models and extract activations from specific layers $l$. Let $\mathbf{A}_1 \in \mathbb{R}^{n \times p_1}$ and $\mathbf{A}_2 \in \mathbb{R}^{n \times p_2}$ denote the activation matrices from the $l$-th layer of the two models, where $p_1$ and $p_2$ are the dimensionalities of the feature representations at that layer.

Next, the activation matrices $\mathbf{A}_1$ and $\mathbf{A}_2$ are transformed into kernel matrices $\mathbf{K}_1$ and $\mathbf{K}_2$ using a kernel function, typically the linear kernel:

\begin{equation} \centering \label{eq_kernel} {\mathbf{K}}= \mathbf{H}({\mathbf{A}} {\mathbf{A}}^{T})\mathbf{H}^T, \end{equation}
where $\mathbf{H} = \mathbf{I}- \frac{1}{n}\textbf{11}^T$ is the centering matrix, with $\mathbf{I}$ as the identity matrix and $\textbf{1}$ as a vector of ones. This transformation ensures that the kernel matrix $\mathbf{K} \in \mathbb{R}^{n \times n}$ eliminates biases introduced by differences in model architecture.

The CKA similarity between the feature representations of two models is then defined as:

\begin{equation} 
\centering \label{eq_cka} 
\text{CKA}(f^1, f^2,\mathbf{X}) = \frac{\text{tr}(\mathbf{K}_1 \mathbf{K}_2)}{\sqrt{\|\mathbf{K}_1\|_F^2 \cdot \|\mathbf{K}_2\|_F^2}},
\end{equation}
where \(\mathbf{K}_1\) and \(\mathbf{K}_2\) are the kernel matrices derived from the activations of models \(f^1\) and \(f^2\) for input \(\mathbf{X}\). The term \(\|\mathbf{K}_*\|_F^2\) represents the squared Frobenius norm, which is computed as the trace of the matrix product \(\mathbf{K}_* \mathbf{K}_*\), \emph{i.e.}, \(\|\mathbf{K}_*\|_F^2 = \text{tr}(\mathbf{K}_* \mathbf{K}_*)\).

CKA is architecture independent because it doesn't change when certain transformations are applied \cite{cka_2024,cka_ori}. This means that architectural differences don't change how similar two models are when measuring similarity. In particular: 1) \emph{Orthogonal transformation invariance}. CKA remains unchanged under rotations and reflections of the feature space, making it robust to different basis representations. 2) \emph{Isotropic scaling invariance}. Uniform scaling of feature representations does not impact CKA values, ensuring that similarity comparisons are not biased by differences in activation magnitudes. Because of these features, CKA is a very good way to compare models with different architectures because it looks at the relative structure of feature representations instead of their absolute values or specific network configurations.

\section{Method}
\begin{figure*}[htbp]
\vspace{-0.3cm}
    \centering
    \includegraphics[width=0.95
    \textwidth]{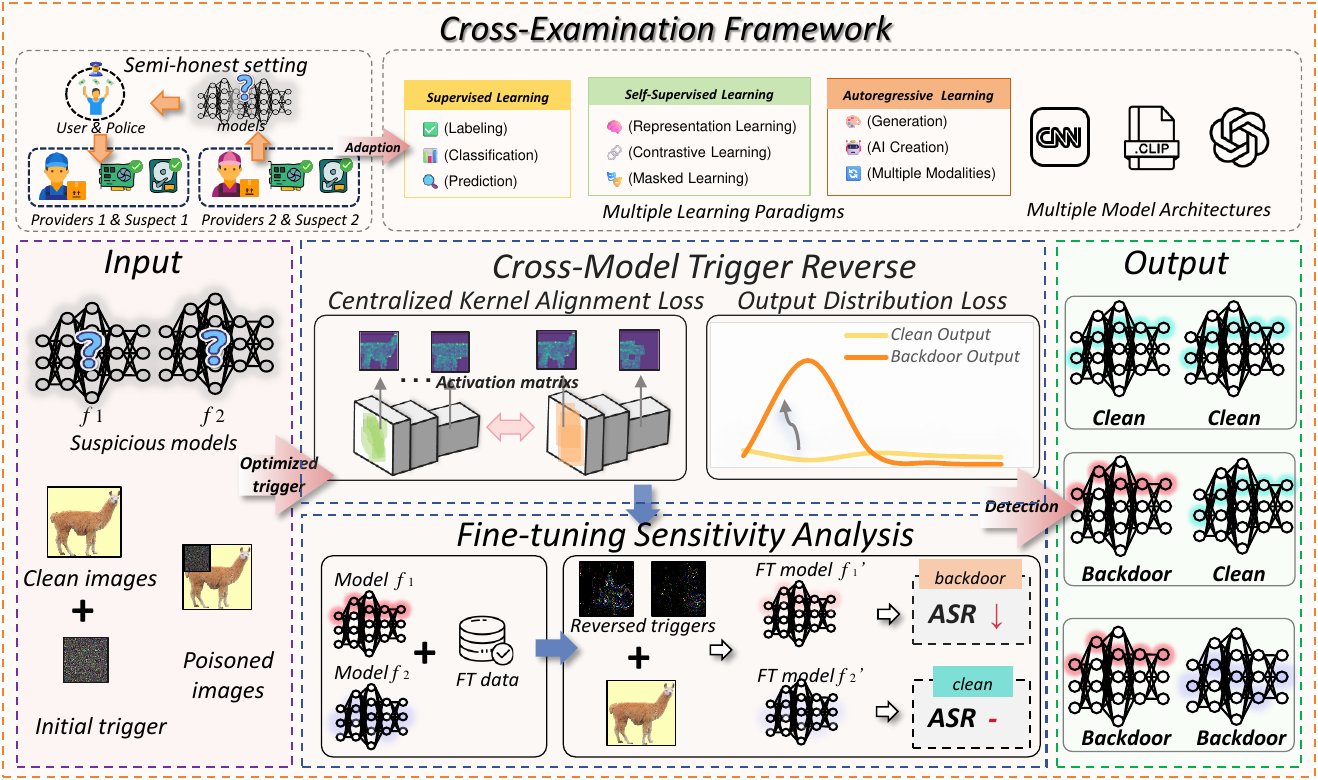} 
    \vspace{-0.3cm}
    \caption{Overview of the Lie Detector. We propose a general backdoor detection method based on the cross-examination framework. By leveraging output distribution loss and CKA loss to reverse triggers and further identifying backdoored models through fine-tuning sensitivity analysis, our approach ensures data security in third-party training processes.} 
    \label{fig:method} 
    \vspace{-0.5cm}
\end{figure*}

In this section, we will introduce backdoor defense method \textbf{Lie Detecor} based on the cross-examination framework as shown in Fig.~\ref{fig:method}.

\subsection{Cross-Examination Framework}
To enhance the security of third-party machine learning models, we propose a \emph{Cross-Examination-Based Backdoor Detection Framework}, designed for a \emph{semi-honest verification setting}. The framework consists of three main modules.

\textbf{Cloud customers}. It consists of users who require model training services but lack direct control over the training process. These users also act as the verification party (the \texttt{police}), who have the authority to verify model integrity. The users provide the clean dataset \( \mathcal{D}_{\text{c}} \), training hyperparameters, model architecture \( f \), and the learning paradigm \( \mathcal{L}_\textit{learn} \). 

\textbf{Semi-honest third-party providers}. They are independent service providers (the \texttt{suspects}) responsible for model training. While they follow the training protocol prescribed by users, they still retain the possibility of embedding arbitrary backdoors into the model. Their malicious behavior is reflected in a data poisoning process, where a fraction of the training data is modified to implant hidden vulnerabilities.

Specifically, we assume an adversary (\texttt{suspect}) trains a model \( f_{\mathbf{\theta}} \) using an original dataset \( D_c \) and alters a subset of it to create poisoned samples \( D_p \) through predefined training details. The poisoned dataset consists of \( \alpha |D_c| \) modified samples, where \( \alpha \in [0,1] \) denotes the poisoning rate. The overall dataset used for training is then:
\begin{equation}
\mathcal{D} = (D_c \setminus D_p) \cup D_p.
\end{equation}

The adversary's learning process can be formulated as an optimization problem:
\begin{equation}
\begin{aligned}
    \arg \min_{\mathbf{\theta}^*} 
    \big\{ 
    &\mathcal{L}_{\text{learn}}(f_{\mathbf{\theta}}, \mathcal{D})
    \triangleq (1 - \alpha) \mathbb{E}_{(\mathbf{x}_c, y_c) \sim D_c} 
    \big[ \ell (f_{\mathbf{\theta}}(\mathbf{x}_c), y_c) \big] \\
    &\quad {} + \alpha \mathbb{E}_{(\mathbf{x}_p, \hat{y}_c) \sim D_p} 
    \big[ \ell (f_{\mathbf{\theta}}(\mathbf{x}_p), \hat{y}_c) \big]
    \big\},
\end{aligned}
\end{equation}
where \( y_c \) is the ground-truth label of a clean sample \( \mathbf{x}_c \), while \( \hat{y}_c \) is the adversarially assigned target label for a poisoned sample \( \mathbf{x}_p \), used to induce backdoor behavior. The learning objective \( L_{\text{learn}} \) varies by learning paradigm, with the loss function \( \ell(\cdot,\cdot) \) defined as follows: In \emph{supervised learning}, \( y \) represents discrete class labels for classification tasks, and \( \ell \) is typically the cross-entropy loss. In \emph{contrastive learning} (\emph{e.g.}, CLIP), \( y \) defines similarity relationships rather than explicit class labels, and \( \ell \) is a similarity-based contrastive loss. In \emph{autoregressive learning} (\emph{e.g.}, LLaVA), \( y \) serves as a reconstruction target, and \( \ell \) includes reconstruction or autoregressive losses.

\textbf{Cross-examination backdoor detection}. The goal of cross-examination backdoor detection is to verify whether a model has been compromised by a backdoor without requiring access to its training data or process. Instead of relying on a known clean reference model or predefined attack patterns, our approach detects backdoors by leveraging inconsistencies between two independently trained models (\( f_1 \) and \( f_2 \)) provided by different third-party service providers. Under the Cross-Examination framework, there are three possible outcomes: both models are clean, both models are backdoored, or one model is clean while the other is backdoored. 

Next we present the challenges and motivations in our framework. \\
\underline{Challenges.} Backdoor detection faces two primary challenges: 1) Accuracy. Many existing detection methods rely heavily on statistical analysis, assuming access to a clean reference model and predefined attack patterns. These assumptions introduce limitations, as mismatched priors can lead to detection failures when facing unknown or adaptive backdoor attacks. Traditional statistical approaches struggle to generalize beyond known attack distributions, reducing their reliability in real-world scenarios. 2) Generalization. The detection framework must be robust across different model architectures and learning paradigms, not just classification tasks. A method that is tightly coupled to specific model types or training objectives may fail in diverse applications, such as semi supervised learning or generative models. Ensuring architectural and task-agnostic generalization is critical for practical deployment.

\underline{Motivations.} Compared to existing backdoor detection methods, our framework introduces the following innovations: 1) Leveraging model inconsistencies to avoid predefining attacks. Traditional detection methods depend on statistical assumptions about the distribution of backdoor triggers or poisoned data, making them vulnerable to novel or adaptive attacks. Our framework circumvents this limitation by exploiting inconsistencies between independently trained models on the same dataset, allowing detection without relying on prior knowledge of attack patterns.
2) Utilizing invariant features for better generalization. Many conventional defenses are tightly coupled to specific model architectures or training paradigms, limiting their applicability beyond classification tasks. Our method focuses on detecting structural inconsistencies that remain invariant across different architectures and learning paradigms, enabling broader applicability in different learning paradigms.

\subsection{Cross-Model Trigger Reverse}
In this subsection, we need to leverage the behavioral differences between models $f_1$ and $f_2$ to detect potential backdoors, conducting an initial screening to identify suspected backdoors in the models.

We generate triggers that effectively activate backdoor behaviors across different learning paradigms, formulating them as a combination of two trainable components: a mask \( \mathbf{m} \) and a pattern \( \mathbf{p} \). The mask \( \mathbf{m} \) controls which pixels in the input image are modified, while the pattern \( \mathbf{p} \) defines the injected adversarial content.  
\begin{equation}
\label{eq_trigger}
     \mathbf{x}'= \mathbf{m}\odot \mathbf{p} + (1-\mathbf{m}) \odot \mathbf{x},
\end{equation}
where \( \mathbf{x} \) and \( \mathbf{x}' \) represent the clean and poisoned inputs, respectively, and \( \odot \) denotes element-wise multiplication. By optimizing \( \mathbf{m} \) and \( \mathbf{p} \), we reconstruct effective triggers that are capable of eliciting malicious behavior in the suspect model.

\textbf{Output distribution loss.} First, we aim to identify a backdoor trigger (\texttt{evidence}) that can effectively activate the compromised model’s hidden behavior. This is achieved by leveraging the output distribution loss, which exploits the inherent characteristics of backdoor traces within a model. Attackers implant backdoors to ensure that the model’s output distribution strongly favors the attack target when the trigger is present, while a clean model exhibits a more uniform output distribution. The output distribution loss is defined as: 
\begin{equation}
\mathcal{L}_{\text{OD}} = \frac{1}{N} \sum_{i=1}^{N} 
\begin{cases}
-\ell_{\text{CE}}(f(\mathbf{x}'_i), y_i), & \text{if SL,} \\[1mm]
\ell_{\text{Sim}}(f(\mathbf{x}'_i), f(y_i)), & \text{if SSL,} \\[1mm]
\mathbb{E}_{(\mathbf{m}, \mathbf{p})} [\sum_{t} \ell_{\text{AR}}(f(\mathbf{x}', \theta)^{(t)}, \hat{y}_i^{(t)})], & \text{if AL.}
\end{cases}
\end{equation}
where \( N \) represents the number of selected samples drawn from the clean dataset \( D_c \). The loss is computed over this subset rather than the full dataset and about 1000 samples. 

Each term in the summation corresponds to a different learning paradigm: 1) Supervised learning (SL). \( \ell_{\text{CE}} \) is the cross-entropy loss, ensuring the backdoored input \( \mathbf{x}'_i \) is classified as the target label \( y_i \). 2) Self-supervised learning (SSL). \( \ell_{\text{Sim}} \) is the similarity loss, measuring how close the feature representations of \( f(\mathbf{x}'_i) \) and \( f(y_i) \) are. Increase dissimilarity to misalign the poisoned representation with clean semantic features. 3) Autoregressive learning (AL). \( t \) is the index over generated tokens in an autoregressive model. \(\hat{y}_i^{(t)} \) is the attacker-defined target output at timestep \( t \), enforcing supervised sequence control over the backdoored generation. And $\ell_{\text{AR}}$ is the autoregressive loss.

\textbf{CKA loss.} To further expose backdoors, we leverage CKA loss to amplify training inconsistencies. Since CKA reflects representation learning objectives, a backdoored model optimizing for both clean and poisoned objectives inevitably diverges from a clean model. By maximizing this divergence, we highlight the \texttt{Lie} hidden within a suspect model.

\begin{theorem} \textbf{(Task-Driven Representational Similarity Theorem)}
\label{eq:task-cka}
Let $f_1$ and $f_2$ be two independently trained models on the same dataset but potentially with different objectives or architectures. The representational similarity between the models, measured by Centered Kernel Alignment (CKA), strongly correlates with their task alignment:

\begin{equation} \rho_{\text{task}}(f_1, f_2) \propto \text{CKA}(\Phi_{f_1}, \Phi_{f_2}), \end{equation} where $\Phi_{f_1}$ and $\Phi_{f_2}$ are feature representations extracted from the models, and $\rho_{\text{task}}$ quantifies their consistency in downstream task performance. Higher CKA similarity implies stronger alignment in decision boundaries and behavior across datasets. 

\end{theorem}

By Theorem~\ref{eq:task-cka}, CKA serves as a reliable metric to assess the alignment of learned representations between models trained under different paradigms. Since backdoored models are optimized for both clean and adversarial objectives, their representations deviate significantly from clean models. We exploit this by computing the CKA similarity between two models on backdoored inputs.

For models $f_1$ and $f_2$, we compute CKA on activation maps extracted from an input $\mathbf{x}'$. The CKA loss is as follows:

\begin{equation} 
\label{eq_cka_loss} 
\mathcal{L}_{\text{CKA}}(\mathbf{K}',l) = 1 - \frac{\text{tr}(\mathbf{K}_1^l(\mathbf{x}') \mathbf{K}_2^l(\mathbf{x}'))}{\sqrt{\|\mathbf{K}_1^l(\mathbf{x}')\|_F^2 \cdot \|\mathbf{K}_2^l(\mathbf{x}')\|_F^2} }, 
\end{equation} where $\mathbf{K}_1^l(\mathbf{x}')$ and $\mathbf{K}_2^l(\mathbf{x}')$ are kernel matrices computed from the activations of models $f_1$ and $f_2$ on the backdoored input $\mathbf{x}'$. By maximizing $\mathcal{L}_{\text{CKA}}$, we construct inputs that accentuate behavioral discrepancies between the models, thereby facilitating the reverse of backdoor triggers.

Finally, we can minimize the above trigger optimization function as shown in \cref{eq_loss}:
\begin{equation}
\centering
\label{eq_loss}
\mathcal{L}(\mathbf{m}, \mathbf{p}) = \alpha \cdot \mathcal{L}_{\text{CKA}} 
   + \beta \cdot \mathcal{L}_{\text{OD}} 
   + \lambda \cdot (\|\mathbf{m}\|_1 + \|\mathbf{p}\|_1),
\end{equation}
where the \(\text{L}_1\) norm for regularization enhances the optimization and learning process of the trigger by promoting sparsity and minimal perturbation, following the principles outlined in the paper DECREE.

After this stage, we can filter out cases where both models are clean and identify cases where at least one model has a backdoor.

\subsection{Fine-tuning Sensitivity Analysis}

To further distinguish true backdoor models from clean models exhibiting unexpected behavior, we conduct a fine-tuning sensitivity analysis. This process helps accurately locate backdoor-implanted models by evaluating their stability under additional training.

\textbf{Fine-tuning setup.} We fine-tune both models, $f_1$ and $f_2$, obtaining fine-tuned versions $f_1'$ and $f_2'$. The fine-tuning process is performed on a subset ($10\%$) of the clean dataset $\mathcal{D}_c$, denoted as $\mathcal{D}_{\text{ft}} \subset \mathcal{D}_c$, by optimizing the learning paradigm-dependent loss $\mathcal{L}_{\text{learn}}$:

\begin{equation}\textstyle
f' = \arg \min_{f} \mathcal{L}_{\text{learn}}(f, \mathcal{D}_{\text{ft}}).
\end{equation}

\textbf{Backdoor identification criterion.} We evaluate the model’s backdoor robustness by measuring the Attack Success Rate (ASR) before and after fine-tuning:

\begin{equation}
\text{ASR}(f') = \mathbb{E}_{\mathbf{x}' \in \mathcal{D}_{\text{b}}} \left[ \mathbb{I}(f'(\mathbf{x}') = \hat{y}_c) \right],
\end{equation}

where $\mathcal{D}_\text{b}$ contains backdoor-embedded inputs $\mathbf{x}'$, $\hat{y}_c$ is the target label, and $\mathbb{I}(\cdot)$ is the indicator function.

A model is flagged as backdoored if fine-tuning reduces ASR by more than $20\%$:

\begin{equation}
\text{Backdoored} \iff \text{ASR}(f) - \text{ASR}(f') > 0.2.
\end{equation}

Since fine-tuning on clean data weakens backdoor effects, a significant ASR drop indicates reliance on the implanted backdoor, confirming its presence.

\section{Experiments}
\begin{table*}[t]
\vspace{-0.5cm}
\renewcommand{\arraystretch}{0.8} 
\setlength{\tabcolsep}{10pt} 
\centering
\small

\caption{Detection accuracies (\%) on ResNet-18. For each attack, we evaluate 20 clean and 20 backdoored models. Detection Success Rate (DSR) and False Positive Rate (FPR) are reported. Bold indicates the best result, and underline indicates the second-best result.}
\vspace{-0.3cm}
\resizebox{\textwidth}{!}{%
\begin{tabular}{llcccccccccccccc}
\toprule
\multirow{2}{*}{\textbf{Dataset}}  & \multirow{2}{*}{\textbf{Attack}} & \multicolumn{2}{c}{\textbf{NC}} & \multicolumn{2}{c}{\textbf{ABS}} & \multicolumn{2}{c}{\textbf{NAD}} & \multicolumn{2}{c}{\textbf{TED}} & \multicolumn{2}{c}{\textbf{MM-BD}} & \multicolumn{2}{c}{\textbf{DECREE}} & \multicolumn{2}{c}{\textbf{Lie Detector}} \\
\cmidrule(lr){3-4} \cmidrule(lr){5-6} \cmidrule(lr){7-8} \cmidrule(lr){9-10} \cmidrule(lr){11-12} \cmidrule(lr){13-14} \cmidrule(lr){15-16}
&& DSR & FPR & DSR & FPR & DSR & FPR & DSR & FPR & DSR & FPR & DSR & FPR & DSR & FPR \\
\midrule
\multirow{3}{*}{CIFAR10} 
& BadNet  & 87.5 & 10.0 & 90.0 & 5.0 & 92.5 & 15.0 & \textbf{100} & \textbf{0.0} & \textbf{100} & \textbf{0.0} & \underline{97.5} & \textbf{0.0} & \textbf{100} & \textbf{0.0} \\
& Blended & 30.0 & 25.0 & 80.0 & 15.0 & 67.5 & 10.0 & 95.0 & 5.0 & \underline{97.5} & \textbf{0.0} & 92.5 & 5.0 & \textbf{100} & \textbf{0.0} \\
& ISSBA   & 25.0 & 30.0 & 37.5 & 40.0 & 50.0 & 20.0 & \underline{95.0} & 10.0 & 92.5 & 5.0 & 90.0 & 5.0 & \textbf{100} & \textbf{0.0} \\
\midrule
\multirow{3}{*}{TinyImgNet}
& BadNet  & 77.5 & 10.0 & 80.0 & 10.0 & 82.5 & 20.0 & 92.5 & 5.0 & 95.0 & \textbf{0.0} & \underline{97.5} & \textbf{0.0} & \textbf{100} & \textbf{0.0} \\
& Blended & 15.0 & 30.0 & 70.0 & 20.0 & 42.5 & 15.0 & 87.5 & 10.0 & 92.5 & 5.0 & \underline{95.0} & \textbf{0.0} & \textbf{100} & \textbf{0.0} \\
& ISSBA   & 10.0 & 40.0 & 25.0 & 25.0 & 40.0 & 25.0 & 90.0 & 10.0 & \underline{95.0} & 5.0 & 92.5 & 10.0 & \textbf{97.5} & \underline{5.0} \\
\bottomrule
\end{tabular}}
\label{tab:defense}
\vspace{-0.5cm}
\end{table*}

\begin{figure}[t]
    \centering
    \begin{subfigure}{0.23\textwidth}
        \centering
        \includegraphics[width=\linewidth]{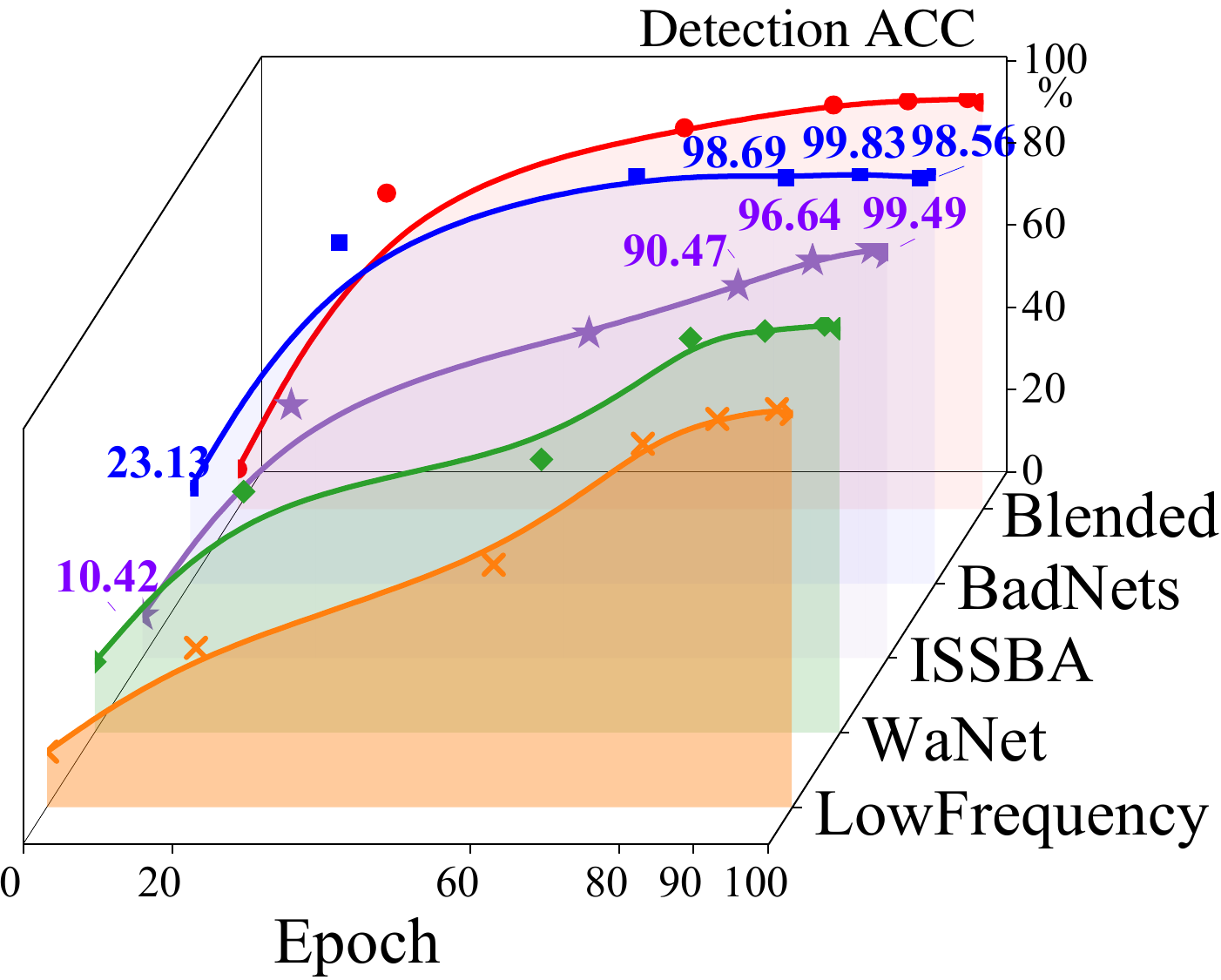}
        \caption{Resnet18}
        \label{fig:rs18}
    \end{subfigure}
    \hfill
    \begin{subfigure}{0.23\textwidth}
        \centering
        \includegraphics[width=\linewidth]{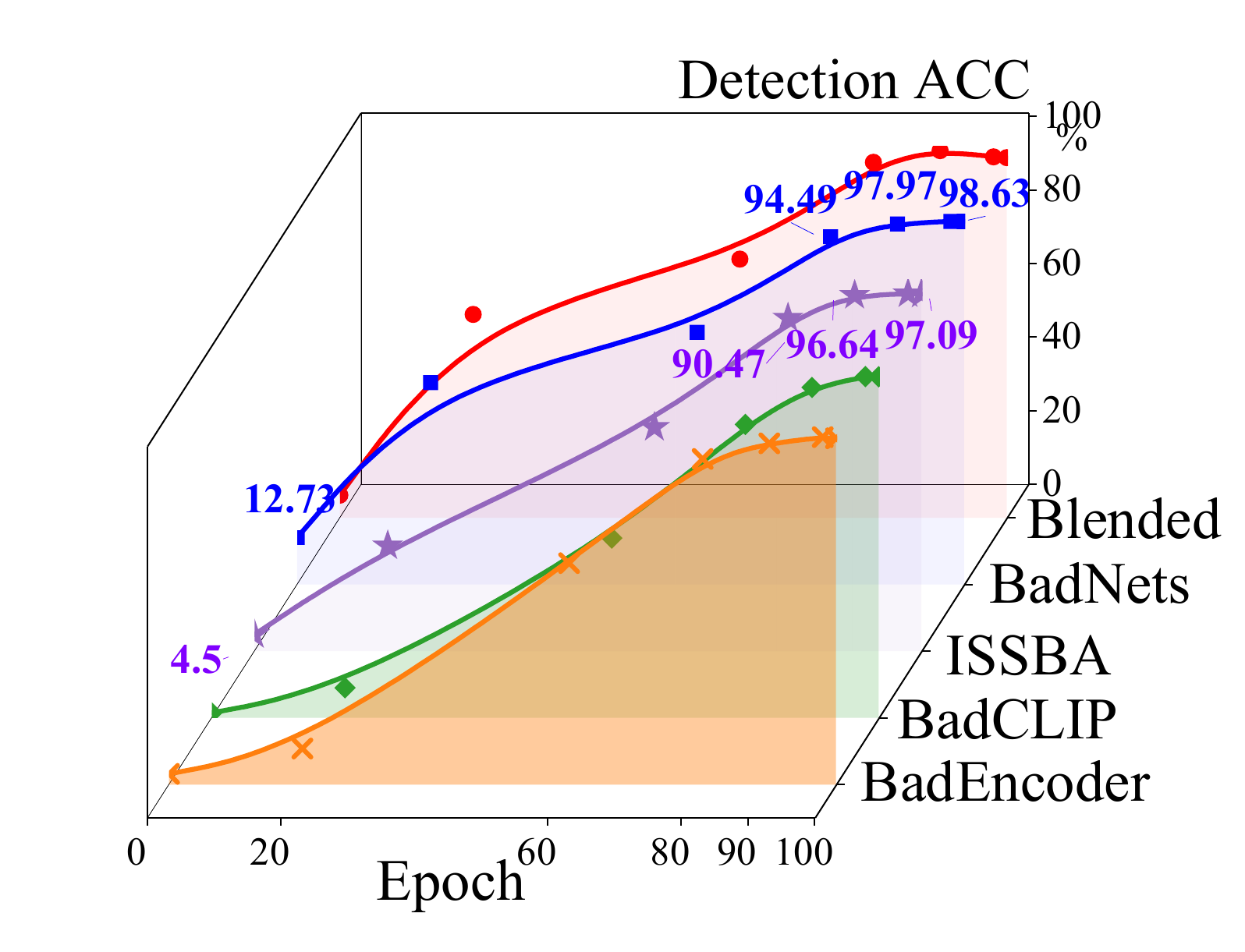}
        \caption{CLIP}
        \label{fig:clip}
    \end{subfigure}
    \vspace{-0.3cm}
    \caption{Detection accuracies of cross-model trigger reverse on ResNet-18 and CLIP}
    \vspace{-0.5cm}
    \label{fig:cka_asr}
\end{figure}

\subsection{Implementation Details}
\textbf{Models and Datasets.} We evaluate our method across multiple learning paradigms. For \emph{supervised learning}, we use ResNet18~\cite{Resnet_2016} and VGG16~\cite{VGG_2015} on CIFAR-10~\cite{cifar} and TinyImageNet~\cite{miniimagenet}. For \emph{self-supervised and autoregressive learning}, we test CLIP~\cite{clip_2021} and CoCoOp on ImageNet~\cite{imagenet} and Caltech101~\cite{caltech101_2004}, while LLaVA~\cite{llava_2023} and mini-GPT-4~\cite{gpt4_2023} are evaluated on COCO~\cite{coco}, Frisk-30k~\cite{f30}, and Frisk-8k~\cite{f8}.

\textbf{Attacks and Defenses.} We consider backdoor attacks across different paradigms, including BadNets~\cite{badnets_2017}, Blended~\cite{blended_2017}, ISSBA~\cite{ISSBA_2021}, WaNet~\cite{wanet_2021}, and Low-Frequency~\cite{low_2021} for \emph{supervised learning}. For \emph{self-supervised and autoregressive learning}, we adapt these attacks and further evaluate BadCLIP~\cite{badclip_2024}, BadEncoder~\cite{badencoder_2022}, TrojanVLM~\cite{trojanvlm_2024}, and Shadowcast~\cite{sw_2024}. We employ advanced defenses, including NC~\cite{NC_2019}, ABS~\cite{ABS_2019}, NAD~\cite{NAD_2021}, TED~\cite{TED_2024}, MM-BD~\cite{MM-BD_2024}, DECREE~\cite{DECREE_2023}, and SEER~\cite{seer_2024}. Some methods, such as TED and MM-BD, are extended to multiple paradigms. Unless otherwise specified, all attack methods use a 10\% poisoning rate. All evaluations are conducted using the semi-honest environment, with detailed settings and evaluation metrics provided in Appendices C.1 and C.2.

\subsection{Detection Performance in SL}
In Tab.~\ref{tab:defense}, we compare our method with six state-of-the-art post-training detection approaches in terms of detection accuracy~\cite{NC_2019,ABS_2019,NAD_2021,TED_2024,MM-BD_2024,DECREE_2023}: NC, ABS, NAD, TED, MM-BD, and DECREE. To evaluate their effectiveness, we first assess these methods on ResNet18 in a supervised learning setup, testing their performance against three classic backdoor attacks on CIFAR-10 and TinyImageNet. We can conclude that: 1) Lie Detector achieves state-of-the-art backdoor detection performance with consistently 100\% DSR and near-zero FPR across different attacks and datasets. This demonstrates its robustness in identifying backdoored models without misclassifying clean ones. 2) Existing detection methods struggle with adaptive backdoor attacks, especially on complex datasets (TinyImageNet). While approaches like TED, MM-BD, and DECREE show improved performance over earlier methods (NC, ABS, NAD), they still fall short in consistently detecting stealthy backdoors (ISSBA).

\subsection{Detection Performance in SSL and AL}

\begin{table*}[t]
\vspace{-0.5cm}
\renewcommand{\arraystretch}{0.8} 
\setlength{\tabcolsep}{12pt} 
\centering
\small
\caption{Detection accuracies (\%) in CLIP and LLaVA. we evaluate 10 clean and 10 backdoored models per attack. Detection Success Rate (DSR) and False Positive Rate (FPR) are reported. Bold indicates the best result, and underline indicates the second-best result.}
\vspace{-0.3cm}
\resizebox{\textwidth}{!}{%
\begin{tabular}{cllcccccccccc}
\toprule
\multirow{2}{*}{\textbf{Architecture}} & \multirow{2}{*}{\textbf{Dataset}} & \multirow{2}{*}{\textbf{Attack}} & \multicolumn{2}{c}{\textbf{TED}} & \multicolumn{2}{c}{\textbf{MM-BD}} & \multicolumn{2}{c}{\textbf{DECREE}} & \multicolumn{2}{c}{\textbf{SEER}} & \multicolumn{2}{c}{\textbf{Lie Detector}} \\
\cmidrule(lr){4-5} \cmidrule(lr){6-7} \cmidrule(lr){8-9} \cmidrule(lr){10-11} \cmidrule(lr){12-13}
&&& DSR & FPR & DSR & FPR & DSR & FPR & DSR & FPR & DSR & FPR \\
\midrule
\multirow{6}{*}{CLIP}
& \multirow{3}{*}{Caltech101} 
& BadNet  & 80.0 & 10.0 & 75.0 & 10.0 & \underline{87.0} & 20.0 & \textbf{100.0} & \textbf{0.0} & \textbf{100.0} & \textbf{0.0} \\
&& Blended & 77.5 & 15.0 & 72.5 & 20.0 & \underline{82.5} & 25.0 & \textbf{97.5} & \textbf{0.0} & \textbf{97.5} & \textbf{0.0} \\
&& BadCLIP & 47.5 & 35.0 & 52.5 & 25.0 & \underline{60.0} & 30.0 & \underline{90.0} & \textbf{0.0} & \textbf{95.0} & 5.0 \\
\cmidrule(lr){2-13}
& \multirow{3}{*}{ImageNet}
& BadNet & 60.0 & 15.0 & 67.5 & 10.0 & 72.5 & 10.0 & \textbf{95.0} & \textbf{0.0} & \textbf{95.0} & \textbf{0.0} \\
&& Blended & 57.5 & 20.0 & 65.0 & 15.0 & 75.0 & 15.0 & \underline{90.0} & 5.0 & \textbf{92.5} & \textbf{0.0} \\
&& BadCLIP & 37.5 & 30.0 & 42.5 & 30.0 & 45.0 & 20.0 & \underline{87.5} & 10.0 & \textbf{90.0} & 5.0 \\
\midrule
\multirow{4}{*}{LLaVA}
& \multirow{2}{*}{COCO} 
& TrojanVLM & 10.0 & 50.0 & 15.0 & 45.0 & 60.0 & 40.0 & \underline{80.0} & 15.0 & \textbf{95.0} & 5.0 \\
&& Shadowcast & 10.0 & 50.0 & 15.0 & 50.0 & 60.0 & 45.0 & \underline{85.0} & 10.0 & \textbf{92.5} & \textbf{0.0} \\
\cmidrule(lr){2-13}
& \multirow{2}{*}{Flickr-30K}
& TrojanVLM & 10.0 & 55.0 & 15.0 & 50.0 & 55.0 & 45.0 & \underline{80.0} & 5.0 & \textbf{90.0} & 10.0 \\
&& Shadowcast & 10.0 & 50.0 & 10.0 & 45.0 & 45.0 & 35.0 & \underline{80.0} & 10.0 & \textbf{90.0} & 5.0 \\
\bottomrule
\end{tabular}}
\label{tab:defense ssl and al}
\vspace{-0.5cm}
\end{table*}
We evaluate our method against four defense approaches under semi-supervised and autoregressive learning paradigms. We follow the original implementations of these methods with only modest modifications. The detection success rates are tested across four datasets and three classic backdoor attacks. Based on Tab.~\ref{tab:defense ssl and al}, we draw the following conclusions: 1) Existing methods have limited generalization. Traditional detection methods (TED, MM-BD, DECREE) show inconsistent performance across datasets and architectures. While some perform well on CLIP, they fail on vision-language models (e.g., LLaVA), indicating weak adaptability across learning paradigms. 2) FPR is High. Many methods, particularly TED and MM-BD, exhibit FPRs as high as 50\%, misclassifying clean models as backdoored at a detection rate no better than random guessing. However, our method achieves superior generalization across different learning paradigms, maintaining high detection success rates with consistently low false positives.




\begin{table}[t]
\renewcommand{\arraystretch}{0.9} 
\setlength{\tabcolsep}{4pt} 
\centering
\small
\caption{Component ablation experiments. }
\vspace{-0.3cm}
\resizebox{0.48\textwidth}{!}{%
\begin{tabular}{llccccc}
\toprule
\textbf{Component} & \textbf{Attack} & \textbf{Task} & \textbf{Trigger Size} & \textbf{Model} & \textbf{DSR} & \textbf{FPR} \\
\midrule
\multirow{6}{*}{\makecell{Cross-Model\\Trigger Reverse}}
& \multirow{2}{*}{Blended} & \multirow{2}{*}{CIFAR10} & \multirow{2}{*}{4×4} & ResNet-18 & 100.0 & 10.0  \\
&&&& VGG16 & 100.0 & 20.0  \\
\cmidrule(lr){2-7}
& \multirow{2}{*}{BadEncoder} & \multirow{2}{*}{Caltech101} & \multirow{2}{*}{32×32} & CLIP & 100.0 & 20.0 \\
&&&& CoCoOp & 100.0 & 20.0  \\
\cmidrule(lr){2-7}
& \multirow{2}{*}{Shadowcast} & \multirow{2}{*}{Flickr8k} & \multirow{2}{*}{50×50} & LLaVA & 90.0 & 20.0  \\
&&&& Mini-GPT4 & 90.0 & 30.0  \\
\midrule
\multirow{6}{*}{\makecell{Lie Detector}}
& \multirow{2}{*}{Blended} & \multirow{2}{*}{CIFAR10} & \multirow{2}{*}{4×4} & ResNet-18 & 100.0 & 0.0  \\
&&&& VGG16 & 100.0 & 0.0  \\
\cmidrule(lr){2-7}
& \multirow{2}{*}{BadEncoder} & \multirow{2}{*}{Caltech101} & \multirow{2}{*}{32×32} & CLIP & 100.0 & 0.0 \\
&&&& CoCoOp & 100.0 & 0.0  \\
\cmidrule(lr){2-7}
& \multirow{2}{*}{Shadowcast} & \multirow{2}{*}{Flickr8k} & \multirow{2}{*}{50×50} & LLaVA & 90.0 & 0.0  \\
&&&& Mini-GPT4 & 90.0 & 0.0  \\
\bottomrule
\end{tabular}}
\label{tab:ablation}
\vspace{-0.3cm}
\end{table}

\subsection{Ablation Study}

To validate the effectiveness of Cross-Model Trigger Reverse and the robustness check phase in the Lie Detector, we conduct component ablation experiments in Tab.~\ref{tab:ablation}. Specifically, the Cross-Model Trigger Reverse setup removes the fine-tuning robustness analysis component, while the Lie Detector includes both components. We can conclude the following: 1) ``Cross-Model Trigger Reverse'' alone is effective but less robust. While it achieves high DSR, its FPR remains relatively high, reaching up to 30\% in some cases, indicating potential misclassifications. 2) ``Fine-tuning Sensitivity Analysis'' significantly improves robustness. By integrating fine-tuning robustness analysis, the Lie Detector maintains the same high DSR while reducing FPR to 0\% across all tested settings, demonstrating its effectiveness in distinguishing backdoored models from clean ones.

\textbf{Similarity metrics selection.} We select four existing similarity metrics for comparative testing, including CKA, CCA (Canonical Correlation Analysis), SVCCA (Singular Vector Canonical Correlation Analysis) and COS (Cosine Similarity)~\cite{lahitani2016cosine}. We adaptively replace the aforementioned different similarity metrics for backdoor detection. The results indicate that CKA outperforms other similarity metrics across different learning paradigms, especially on the LLaVA model, which also indirectly demonstrates the architecture-agnostic nature of the CKA metric.

\textbf{Number of epochs.} We present the detection accuracies under DSR as the number of epochs increases for ResNet-18 and CLIP models, as shown in Fig.~\ref{fig:cka_asr}. We observe that our method achieves stable convergence and remains effective across all attack methods on both ResNet-18 and CLIP models. The detection accuracy consistently improves with training epochs, demonstrating the robustness and adaptability of our approach in identifying backdoored models across different architectures and learning paradigms.

\begin{table}[t]
\renewcommand{\arraystretch}{0.85} 
\setlength{\tabcolsep}{4pt} 
\centering
\small
\caption{Detection accuracies of methods with different model architectures.}
\vspace{-0.3cm}
\resizebox{0.48\textwidth}{!}{%
\begin{tabular}{llccccc}
\toprule
\textbf{Attack} & \textbf{Task} & \textbf{Trigger Size} & \textbf{Model} & \textbf{DSR} & \textbf{FPR} & \textbf{FLOPs} \\
\midrule
\multirow{2}{*}{BadNet} & \multirow{2}{*}{CIFAR10} & \multirow{2}{*}{4×4} & ResNet-18 & 100.0 & 0.0 & 0.7 \\
&&& VGG16 & 100.0 & 0.0 & 0.4 \\
\midrule
\multirow{2}{*}{BadCLIP} & \multirow{2}{*}{Caltech101} & \multirow{2}{*}{32×32} & CLIP & 90.0 & 0.0 & 4.9 \\
&&& CoCoOp & 100.0 & 0.0 & 5.0 \\
\midrule
\multirow{2}{*}{TrojanVLM} & \multirow{2}{*}{Flickr8k} & \multirow{2}{*}{50×50} & LLaVA & 90.0 & 0.0 & 76.6 \\
&&& Mini-GPT4 & 80.0 & 0.0 & 80.3 \\
\bottomrule
\end{tabular}}
\label{tab:architecture}
\vspace{-0.6cm}
\end{table}

\textbf{Model architecture}. We evaluate the effectiveness of our detection method across six different model architectures spanning three learning paradigms, as shown in Table~\ref{tab:architecture}. Specifically, we assess supervised learning models (ResNet-18, VGG16), contrastive language-image models (CLIP, CoCoOp), and vision-language models (LLaVA, Mini-GPT4) under three representative backdoor attacks. We can conclude that: 1) Our method achieves 100\% DSR and 0\% FPR across diverse architectures, including complex multimodal models like LLaVA and Mini-GPT4. 2) Detection performance remains unaffected by model complexity, as measured by FLOPs. For example, despite a significant increase in computational cost from ResNet-18 (0.7 GFLOPs) to Mini-GPT4 (80.3 GFLOPs), our method consistently delivers high DSR with zero false positives.

\begin{table}[t]
\renewcommand{\arraystretch}{0.85} 
\setlength{\tabcolsep}{14pt} 
\centering
\small
\caption{Variation of CKA values under different layers.}
\vspace{-0.3cm}
\resizebox{0.48\textwidth}{!}{%
\begin{tabular}{lcccc}
\toprule
\multirow{2}{*}{\textbf{Layer}} & \multicolumn{2}{c}{\textbf{ResNet-18}} & \multicolumn{2}{c}{\textbf{CLIP}} \\
\cmidrule(lr){2-3} \cmidrule(lr){4-5}
 & Clean & Backdoor & Clean & Backdoor \\
\midrule
layer1 & 0.974 & 0.945 & 0.891 & 0.863 \\
layer2 & 0.936 & 0.768 & 0.853 & 0.632 \\
layer3 & 0.901 & 0.542 & 0.810 & 0.497 \\
layer4 & 0.872 & \textbf{0.427} & 0.795 & \textbf{0.314} \\
\bottomrule
\end{tabular}}
\label{tab:cka_probe}
\vspace{-0.55cm}
\end{table}

\textbf{Feature layer selection}. As shown in Tab.~\ref{tab:cka_probe}, CKA values in clean models remain stable across layers, whereas backdoored models exhibit a notable decline in deeper layers. This trend is consistent across ResNet-18 (supervised learning) and CLIP (SSL), confirming CKA’s reliability as a backdoor probe. Notably, layer 4 yields the most significant CKA drop (0.427 in ResNet-18, 0.314 in CLIP), making it the most effective layer for detection. A possible reason is that higher-layer features capture more abstract semantic information, which backdoor triggers distort, leading to greater representation shifts. So we choose the fourth layer features to calculate the CKA loss.


\section{Conclusion}

This paper proposes a unified backdoor detection framework for semi-honest settings where model training is outsourced to third-party providers. By leveraging cross-examination of model inconsistencies between independent service providers, our method significantly improves detection robustness across different learning paradigms. We integrate Centered Kernel Alignment (CKA) for precise feature similarity measurement and fine-tuning sensitivity analysis to distinguish backdoor triggers from adversarial perturbations, effectively reducing false positives. Extensive experiments demonstrate that our approach outperforms state-of-the-art methods, achieving superior detection accuracy in supervised, contrastive, and autoregressive learning tasks. Notably, it is the first to effectively detect backdoors in multimodal large language models. This work provides a practical solution to mitigate backdoor risks in outsourced model training, paving the way for more secure and trustworthy AI systems.

{
    \small
    \bibliographystyle{ieeenat_fullname}
    \bibliography{main}

\begin{thebibliography}{49}
\providecommand{\natexlab}[1]{#1}
\providecommand{\url}[1]{\texttt{#1}}
\expandafter\ifx\csname urlstyle\endcsname\relax
  \providecommand{\doi}[1]{doi: #1}\else
  \providecommand{\doi}{doi: \begingroup \urlstyle{rm}\Url}\fi

\bibitem[Chen et~al.(2020)Chen, Kornblith, Norouzi, and Hinton]{sim_2020}
Ting Chen, Simon Kornblith, Mohammad Norouzi, and Geoffrey Hinton.
\newblock A simple framework for contrastive learning of visual representations.
\newblock In \emph{Proceedings of the 37th International Conference on Machine Learning}, pages 1597--1607. PMLR, 2020.

\bibitem[Chen et~al.(2017)Chen, Liu, Li, Lu, and Song]{blended_2017}
Xinyun Chen, Chang Liu, Bo Li, Kimberly Lu, and Dawn Song.
\newblock Targeted backdoor attacks on deep learning systems using data poisoning, 2017.

\bibitem[Ciernik et~al.(2024)Ciernik, Linhardt, Morik, Dippel, Kornblith, and Muttenthaler]{cka_2024}
Laure Ciernik, Lorenz Linhardt, Marco Morik, Jonas Dippel, Simon Kornblith, and Lukas Muttenthaler.
\newblock Training objective drives the consistency of representational similarity across datasets, 2024.

\bibitem[Dargan et~al.(2020)Dargan, Kumar, Ayyagari, and Kumar]{dargan2020survey}
Shaveta Dargan, Munish Kumar, Maruthi~Rohit Ayyagari, and Gulshan Kumar.
\newblock A survey of deep learning and its applications: a new paradigm to machine learning.
\newblock \emph{Archives of computational methods in engineering}, 27:\penalty0 1071--1092, 2020.

\bibitem[Deng et~al.(2009{\natexlab{a}})Deng, Dong, Socher, Li, Li, and Fei-Fei]{ImageNet_2009}
Jia Deng, Wei Dong, Richard Socher, Li-Jia Li, Kai Li, and Li Fei-Fei.
\newblock Imagenet: A large-scale hierarchical image database.
\newblock In \emph{2009 IEEE Conference on Computer Vision and Pattern Recognition}, pages 248--255, 2009{\natexlab{a}}.

\bibitem[Deng et~al.(2009{\natexlab{b}})Deng, Dong, Socher, Li, Li, and Fei-Fei]{imagenet}
Jia Deng, Wei Dong, Richard Socher, Li-Jia Li, Kai Li, and Li Fei-Fei.
\newblock Imagenet: A large-scale hierarchical image database.
\newblock In \emph{2009 IEEE Conference on Computer Vision and Pattern Recognition}, pages 248--255, 2009{\natexlab{b}}.

\bibitem[Devlin et~al.(2019)Devlin, Chang, Lee, and Toutanova]{bert_2019}
Jacob Devlin, Ming-Wei Chang, Kenton Lee, and Kristina Toutanova.
\newblock {BERT}: Pre-training of deep bidirectional transformers for language understanding.
\newblock In \emph{Proceedings of the 2019 Conference of the North {A}merican Chapter of the Association for Computational Linguistics: Human Language Technologies, Volume 1 (Long and Short Papers)}, pages 4171--4186, Minneapolis, Minnesota, 2019. Association for Computational Linguistics.

\bibitem[Fang et~al.(2015)Fang, Gupta, Iandola, Srivastava, Deng, Dollar, Gao, He, Mitchell, Platt, Lawrence~Zitnick, and Zweig]{f30}
Hao Fang, Saurabh Gupta, Forrest Iandola, Rupesh~K. Srivastava, Li Deng, Piotr Dollar, Jianfeng Gao, Xiaodong He, Margaret Mitchell, John~C. Platt, C. Lawrence~Zitnick, and Geoffrey Zweig.
\newblock From captions to visual concepts and back.
\newblock In \emph{Proceedings of the IEEE Conference on Computer Vision and Pattern Recognition (CVPR)}, 2015.

\bibitem[Fei-Fei et~al.(2004)Fei-Fei, Fergus, and Perona]{caltech101_2004}
Li Fei-Fei, R. Fergus, and P. Perona.
\newblock Learning generative visual models from few training examples: An incremental bayesian approach tested on 101 object categories.
\newblock In \emph{2004 Conference on Computer Vision and Pattern Recognition Workshop}, pages 178--178, 2004.

\bibitem[Feng et~al.(2023)Feng, Tao, Cheng, Shen, Xu, Liu, Zhang, Ma, and Zhang]{DECREE_2023}
Shiwei Feng, Guanhong Tao, Siyuan Cheng, Guangyu Shen, Xiangzhe Xu, Yingqi Liu, Kaiyuan Zhang, Shiqing Ma, and Xiangyu Zhang.
\newblock Detecting backdoors in pre-trained encoders.
\newblock \emph{arXiv}, 2023.

\bibitem[Gu et~al.(2017)Gu, Dolan-Gavitt, and Garg]{badnets_2017}
Tianyu Gu, Brendan Dolan-Gavitt, and Siddharth Garg.
\newblock Badnets: Identifying vulnerabilities in the machine learning model supply chain.
\newblock \emph{Learning}, 2017.

\bibitem[He et~al.(2016)He, Zhang, Ren, and Sun]{Resnet_2016}
Kaiming He, Xiangyu Zhang, Shaoqing Ren, and Jian Sun.
\newblock Deep residual learning for image recognition.
\newblock In \emph{Proceedings of the IEEE Conference on Computer Vision and Pattern Recognition (CVPR)}, 2016.

\bibitem[Hodosh et~al.(2013)Hodosh, Young, and Hockenmaier]{f8}
Micah Hodosh, Peter Young, and Julia Hockenmaier.
\newblock Framing image description as a ranking task: Data, models and evaluation metrics.
\newblock \emph{Journal of Artificial Intelligence Research}, pages 853--899, 2013.

\bibitem[Jia et~al.(2021)Jia, Liu, and Gong]{badencoder_2022}
Jinyuan Jia, Yupei Liu, and Neil~Zhenqiang Gong.
\newblock Badencoder: Backdoor attacks to pre-trained encoders in self-supervised learning, 2021.

\bibitem[Kornblith et~al.(2019)Kornblith, Norouzi, Lee, and Hinton]{cka_ori}
Simon Kornblith, Mohammad Norouzi, Honglak Lee, and Geoffrey Hinton.
\newblock Similarity of neural network representations revisited.
\newblock \emph{Statistics}, 2019.

\bibitem[Krizhevsky~A(2009)]{cifar}
Hinton~G. Krizhevsky~A.
\newblock Learning multiple layers of features from tiny images.
\newblock 2009.

\bibitem[Lahitani et~al.(2016)Lahitani, Permanasari, and Setiawan]{lahitani2016cosine}
Alfirna~Rizqi Lahitani, Adhistya~Erna Permanasari, and Noor~Akhmad Setiawan.
\newblock Cosine similarity to determine similarity measure: Study case in online essay assessment.
\newblock In \emph{2016 4th International conference on cyber and IT service management}, pages 1--6. IEEE, 2016.

\bibitem[LeCun et~al.(1998)LeCun, Bottou, Bengio, and Haffner]{CNN_1998}
Y. LeCun, L. Bottou, Y. Bengio, and P. Haffner.
\newblock Gradient-based learning applied to document recognition.
\newblock \emph{Proceedings of the IEEE}, pages 2278--2324, 1998.

\bibitem[Li et~al.(2023)Li, Wong, Zhang, Usuyama, Liu, Yang, Naumann, Poon, and Gao]{llava_2023}
Chunyuan Li, Cliff Wong, Sheng Zhang, Naoto Usuyama, Haotian Liu, Jianwei Yang, Tristan Naumann, Hoifung Poon, and Jianfeng Gao.
\newblock Llava-med: Training a large language-and-vision assistant for biomedicine in one day.
\newblock In \emph{37th Conference on Neural Information Processing Systems, NeurIPS 2023}, 2023.

\bibitem[Li et~al.(2021{\natexlab{a}})Li, Li, Wu, Li, He, and Lyu]{ISSBA_2021}
Yuezun Li, Yiming Li, Baoyuan Wu, Longkang Li, Ran He, and Siwei Lyu.
\newblock Invisible backdoor attack with sample-specific triggers.
\newblock In \emph{2021 IEEE/CVF International Conference on Computer Vision (ICCV)}, pages 16443--16452, 2021{\natexlab{a}}.

\bibitem[Li et~al.(2021{\natexlab{b}})Li, Lyu, Koren, Lyu, Li, and Ma]{NAD_2021}
Yige Li, Xixiang Lyu, Nodens Koren, Lingjuan Lyu, Bo Li, and Xingjun Ma.
\newblock Neural attention distillation: Erasing backdoor triggers from deep neural networks.
\newblock \emph{CoRR}, abs/2101.05930, 2021{\natexlab{b}}.

\bibitem[Liang et~al.(2024{\natexlab{a}})Liang, Liang, Liu, Jia, Kuang, and Cao]{liang2024poisoned}
Jiawei Liang, Siyuan Liang, Aishan Liu, Xiaojun Jia, Junhao Kuang, and Xiaochun Cao.
\newblock Poisoned forgery face: Towards backdoor attacks on face forgery detection.
\newblock \emph{arXiv preprint arXiv:2402.11473}, 2024{\natexlab{a}}.

\bibitem[Liang et~al.(2024{\natexlab{b}})Liang, Liang, Luo, Liu, Han, Chang, and Cao]{trojanvlm_2024}
Jiawei Liang, Siyuan Liang, Man Luo, Aishan Liu, Dongchen Han, Ee-Chien Chang, and Xiaochun Cao.
\newblock Vl-trojan: Multimodal instruction backdoor attacks against autoregressive visual language models, 2024{\natexlab{b}}.

\bibitem[Liang et~al.(2023)Liang, Zhu, Liu, Wu, Cao, and Chang]{liang2023badclip}
Siyuan Liang, Mingli Zhu, Aishan Liu, Baoyuan Wu, Xiaochun Cao, and Ee-Chien Chang.
\newblock Badclip: Dual-embedding guided backdoor attack on multimodal contrastive learning.
\newblock \emph{arXiv preprint arXiv:2311.12075}, 2023.

\bibitem[Liang et~al.(2024{\natexlab{c}})Liang, Gong, Fang, Liu, Wang, Liu, Cao, Tao, and Ee-Chien]{liang2024red}
Siyuan Liang, Jiajun Gong, Tianmeng Fang, Aishan Liu, Tao Wang, Xianglong Liu, Xiaochun Cao, Dacheng Tao, and Chang Ee-Chien.
\newblock Red pill and blue pill: Controllable website fingerprinting defense via dynamic backdoor learning.
\newblock \emph{arXiv preprint arXiv:2412.11471}, 2024{\natexlab{c}}.

\bibitem[Liang et~al.(2024{\natexlab{d}})Liang, Liang, Pang, Du, Liu, Chang, and Cao]{liang2024revisiting}
Siyuan Liang, Jiawei Liang, Tianyu Pang, Chao Du, Aishan Liu, Ee-Chien Chang, and Xiaochun Cao.
\newblock Revisiting backdoor attacks against large vision-language models.
\newblock \emph{arXiv preprint arXiv:2406.18844}, 2024{\natexlab{d}}.

\bibitem[Liang et~al.(2024{\natexlab{e}})Liang, Zhu, Liu, Wu, Cao, and Chang]{badclip_2024}
Siyuan Liang, Mingli Zhu, Aishan Liu, Baoyuan Wu, Xiaochun Cao, and Ee-Chien Chang.
\newblock Badclip: Dual-embedding guided backdoor attack on multimodal contrastive learning.
\newblock In \emph{Proceedings of the IEEE/CVF Conference on Computer Vision and Pattern Recognition (CVPR)}, pages 24645--24654, 2024{\natexlab{e}}.

\bibitem[Lin et~al.(2014)Lin, Maire, Belongie, Hays, Perona, Ramanan, Doll{\'a}r, and Zitnick]{coco}
Tsung-Yi Lin, Michael Maire, Serge Belongie, James Hays, Pietro Perona, Deva Ramanan, Piotr Doll{\'a}r, and C.~Lawrence Zitnick.
\newblock Microsoft coco: Common objects in context.
\newblock In \emph{Computer Vision -- ECCV 2014}, pages 740--755, Cham, 2014. Springer International Publishing.

\bibitem[Liu et~al.(2023)Liu, Zhang, Xiao, Zhou, Liang, Wang, Liu, Cao, and Tao]{liu2023pre}
Aishan Liu, Xinwei Zhang, Yisong Xiao, Yuguang Zhou, Siyuan Liang, Jiakai Wang, Xianglong Liu, Xiaochun Cao, and Dacheng Tao.
\newblock Pre-trained trojan attacks for visual recognition.
\newblock \emph{arXiv preprint arXiv:2312.15172}, 2023.

\bibitem[Liu et~al.(2024)Liu, Zhou, Liu, Zhang, Liang, Wang, Pu, Li, Zhang, Zhou, et~al.]{liu2024compromising}
Aishan Liu, Yuguang Zhou, Xianglong Liu, Tianyuan Zhang, Siyuan Liang, Jiakai Wang, Yanjun Pu, Tianlin Li, Junqi Zhang, Wenbo Zhou, et~al.
\newblock Compromising embodied agents with contextual backdoor attacks.
\newblock \emph{arXiv preprint arXiv:2408.02882}, 2024.

\bibitem[Liu et~al.(2025)Liu, Liang, Han, Luo, Liu, Cai, He, and Tao]{liu2025elba}
Xuxu Liu, Siyuan Liang, Mengya Han, Yong Luo, Aishan Liu, Xiantao Cai, Zheng He, and Dacheng Tao.
\newblock Elba-bench: An efficient learning backdoor attacks benchmark for large language models.
\newblock \emph{arXiv preprint arXiv:2502.18511}, 2025.

\bibitem[Liu et~al.(2019)Liu, Ma, Lee, Aafer, Tao, and Zhang]{ABS_2019}
Yingqi Liu, Shiqing Ma, Wen-Chuan Lee, Yousra Aafer, Guanhong Tao, and Xiangyu Zhang.
\newblock Abs: Scanning neural networks for back-doors by artificial brain stimulation.
\newblock In \emph{CCS '19: Proceedings of the 2019 ACM SIGSAC Conference on Computer and Communications Security}, 2019.

\bibitem[Mo et~al.(2024)Mo, Zhang, Zhang, Luo, Sun, Hu, Gao, and Xiang]{TED_2024}
Xiaoxing Mo, Yechao Zhang, Leo~Yu Zhang, Wei Luo, Nan Sun, Shengshan Hu, Shang Gao, and Yang Xiang.
\newblock Robust backdoor detection for deep learning via topological evolution dynamics.
\newblock In \emph{2024 IEEE Symposium on Security and Privacy (SP)}, 2024.

\bibitem[Nguyen and Tran(2021)]{wanet_2021}
Tuan~Anh Nguyen and Anh~Tuan Tran.
\newblock Wanet - imperceptible warping-based backdoor attack.
\newblock \emph{CoRR}, abs/2102.10369, 2021.

\bibitem[Radford et~al.(2021)Radford, Kim, Hallacy, Ramesh, Goh, Agarwal, Sastry, Askell, Mishkin, Clark, Krueger, and Sutskever]{clip_2021}
Alec Radford, Jong~Wook Kim, Chris Hallacy, Aditya Ramesh, Gabriel Goh, Sandhini Agarwal, Girish Sastry, Amanda Askell, Pamela Mishkin, Jack Clark, Gretchen Krueger, and Ilya Sutskever.
\newblock Learning transferable visual models from natural language supervision.
\newblock In \emph{Proceedings of the 38th International Conference on Machine Learning}, pages 8748--8763. PMLR, 2021.

\bibitem[Simonyan and Zisserman(2015)]{VGG_2015}
Karen Simonyan and Andrew Zisserman.
\newblock Very deep convolutional networks for large-scale image recognition.
\newblock 2015.

\bibitem[Vinyals et~al.(2016)Vinyals, Blundell, Lillicrap, kavukcuoglu, and Wierstra]{miniimagenet}
Oriol Vinyals, Charles Blundell, Timothy Lillicrap, koray kavukcuoglu, and Daan Wierstra.
\newblock Matching networks for one shot learning.
\newblock In \emph{Advances in Neural Information Processing Systems}. Curran Associates, Inc., 2016.

\bibitem[Wang et~al.(2019)Wang, Yao, Shan, Li, Viswanath, Zheng, and Zhao]{NC_2019}
Bolun Wang, Yuanshun Yao, Shawn Shan, Huiying Li, Bimal Viswanath, Haitao Zheng, and Ben~Y. Zhao.
\newblock Neural cleanse: Identifying and mitigating backdoor attacks in neural networks.
\newblock In \emph{2019 IEEE Symposium on Security and Privacy (SP)}, 2019.

\bibitem[Wang et~al.(2024)Wang, Xiang, Miller, and Kesidis]{MM-BD_2024}
Hang Wang, Zhen Xiang, David~J. Miller, and George Kesidis.
\newblock Mm-bd: Post-training detection of backdoor attacks with arbitrary backdoor pattern types using a maximum margin statistic.
\newblock In \emph{2024 IEEE Symposium on Security and Privacy (SP)}, 2024.

\bibitem[Wang et~al.(2022)Wang, Shi, Min, Wu, Liang, Wu, Liang, and Liu]{wang2022universal}
Yuhang Wang, Huafeng Shi, Rui Min, Ruijia Wu, Siyuan Liang, Yichao Wu, Ding Liang, and Aishan Liu.
\newblock Universal backdoor attacks detection via adaptive adversarial probe.
\newblock \emph{arXiv preprint arXiv:2209.05244}, 2022.

\bibitem[Wang et~al.(2023)Wang, Zhang, Liang, and Wang]{cka_1}
Zhiyuan Wang, Zeliang Zhang, Siyuan Liang, and Xiaosen Wang.
\newblock Diversifying the high-level features for better adversarial transferability.
\newblock Classification performance;Empirical evaluations;Gradient calculations;High-level features;Input transformation;Invariant features;Parameterized;Performance;Real-world;White-box models;, 2023.

\bibitem[Xiao et~al.(2024)Xiao, Liu, Zhang, Zhang, Li, Liang, Liu, Liu, and Tao]{xiao2024bdefects4nn}
Yisong Xiao, Aishan Liu, Xinwei Zhang, Tianyuan Zhang, Tianlin Li, Siyuan Liang, Xianglong Liu, Yang Liu, and Dacheng Tao.
\newblock Bdefects4nn: A backdoor defect database for controlled localization studies in neural networks.
\newblock \emph{arXiv preprint arXiv:2412.00746}, 2024.

\bibitem[Xu et~al.(2024)Xu, Yao, Shu, Sun, Wu, Yu, Goldstein, and Huang]{sw_2024}
Yuancheng Xu, Jiarui Yao, Manli Shu, Yanchao Sun, Zichu Wu, Ning Yu, Tom Goldstein, and Furong Huang.
\newblock Shadowcast: Stealthy data poisoning attacks against vision-language models, 2024.

\bibitem[Zeng et~al.(2021)Zeng, Park, Mao, and Jia]{low_2021}
Yi Zeng, Won Park, Z.~Morley Mao, and Ruoxi Jia.
\newblock Rethinking the backdoor attacks' triggers: A frequency perspective.
\newblock In \emph{Proceedings of the IEEE/CVF International Conference on Computer Vision (ICCV)}, pages 16473--16481, 2021.

\bibitem[Zhang et~al.(2024)Zhang, Liu, Zhang, Liang, and Liu]{zhang2024towards}
Xinwei Zhang, Aishan Liu, Tianyuan Zhang, Siyuan Liang, and Xianglong Liu.
\newblock Towards robust physical-world backdoor attacks on lane detection.
\newblock \emph{arXiv preprint arXiv:2405.05553}, 2024.

\bibitem[Zhou et~al.(2022)Zhou, Yang, Loy, and Liu]{coop_2022}
Kaiyang Zhou, Jingkang Yang, Chen~Change Loy, and Ziwei Liu.
\newblock Conditional prompt learning for vision-language models.
\newblock In \emph{Proceedings of the IEEE/CVF Conference on Computer Vision and Pattern Recognition (CVPR)}, pages 16816--16825, 2022.

\bibitem[Zhu et~al.(2023)Zhu, Chen, Shen, Li, and Elhoseiny]{gpt4_2023}
Deyao Zhu, Jun Chen, Xiaoqian Shen, Xiang Li, and Mohamed Elhoseiny.
\newblock Minigpt-4: Enhancing vision-language understanding with advanced large language models, 2023.

\bibitem[Zhu et~al.(2024{\natexlab{a}})Zhu, Ning, Li, Xin, and Wu]{seer_2024}
Liuwan Zhu, Rui Ning, Jiang Li, Chunsheng Xin, and Hongyi Wu.
\newblock Seer: Backdoor detection for vision-language models through searching target text and image trigger jointly.
\newblock In \emph{Proceedings of the AAAI Conference on Artificial Intelligence (AAAI)}, pages 7766--7774, 2024{\natexlab{a}}.

\bibitem[Zhu et~al.(2024{\natexlab{b}})Zhu, Liang, and Wu]{zhu2024breaking}
Mingli Zhu, Siyuan Liang, and Baoyuan Wu.
\newblock Breaking the false sense of security in backdoor defense through re-activation attack.
\newblock \emph{arXiv preprint arXiv:2405.16134}, 2024{\natexlab{b}}.

\end{thebibliography}
}

\end{document}